\documentclass{edm_article}

\begin{document}

\title{Applying Recent Innovations from NLP to MOOC Student Course Trajectory Modeling}
%
\numberofauthors{2}
\author{
\alignauthor
Clarence Chen\\
       \affaddr{University of California, Berkeley}\\
       \email{clarencechenct@berkeley.edu}
\alignauthor
Zachary Pardos\\
       \affaddr{UC Berkeley School of Information}\\
       \email{zp@ischool.berkeley.edu}
}
\maketitle


\begin{abstract}
  This paper presents several strategies that can improve neural network-based predictive methods for MOOC student course trajectory modeling, applying multiple ideas previously applied to tackle NLP (Natural Language Processing) tasks. In particular, this paper investigates LSTM networks enhanced with two forms of regularization, along with the more recently introduced Transformer architecture.
\end{abstract}

%


\section{Introduction and Model Outlines}
\label{sec:intro}
\subsection{Fundamentals of Predictive Modeling}
Recent innovations in deep learning methods for NLP (Natural Language Processing) tasks such as \cite{Vaswani17} in the past few years have consistently pushed the state of the art in a wide range of benchmark NLP tasks, while yielding new strategies that can be applied to predictive modeling tasks in a more general sense. This is because the majority of these NLP tasks within the scope of these innovations can be parameterized in terms of modeling the function $f$ in the equation
\begin{equation}
    P(y | x_0, \dots x_t) = f(x_0, \dots x_t ; \theta)
\end{equation}
where $f$ is a probability mass function with parameters $\theta$ over the random variable $y$, and $x_0, \dots x_t$, drawn from a discrete set of tokens $T$, represent the context from previous time steps. Unfortunately, there is little literature in the domain of education analytics exploring the effectiveness of innovations from NLP for education analytics tasks that also conform to this predictive modeling paradigm. Nevertheless, the LSTM (Long-Short-Term Memory) DNN (Deep Neural Network) architecture, an earlier innovation which was the architecture of choice for predictive modeling tasks in NLP before the past few years, has been successfully applied to several education analytics tasks. These papers demonstrate potential for further exploitation of the similarities between predictive modeling tasks in education analytics and NLP, while providing a baseline to compare with more recent innovations presented in this paper.
\subsection{Previous Work with MOOC Course Trajectory Modeling}
One of the first papers to present an application of DNN models for predictive modeling tasks in education analytics is \cite{Pardos17}, where the authors specifically investigate the applicability of DNN models for modeling student course trajectories in MOOCs (Massive Open Online Courses). Specifically, the authors of \cite{Pardos17} demonstrate the effectiveness of LSTM DNN models for this task over other strategies, such as using $n$-gram models that condition their predictions over small number of past course nodes. Finally, the authors provides suggestion for incorporating such a predictive model in a wider context, including tie-ins with the MOOC service to provide user-facing suggestions and live feedback to monitor the predictive model's performance.  
\subsection{Baseline LSTM}
This model is identical to the Baseline LSTM model featured in \cite{Pardos17}, using the same LSTM architecture and a nearly identical hyperparameter set and training scheme, further detailed in Section~\ref{sec:hyperparameters}. This model is intended as a control baseline to assess the performance of other models tested in this paper.
\subsection{Transformer Architecture}
\label{sec:transformer}
As noted in Section~\ref{sec:intro}, the shared abstraction of both course trajectory modeling and many NLP tasks as a discrete next-step predictive task suggests applying innovations from NLP to improve performance in course trajectory modeling. In particular, the Transformer architecture, first featured in \cite{Vaswani17}, is one such major architectural innovation.
\paragraph{Transformer Architecture Details}
In \cite{Vaswani17}, the authors construct a DNN model architecture centered around modular Transformer blocks as described in Section~\ref{sec:attention}. In contrast with the need for $O(n)$ forward and backward passes per input sequence through each of the LSTM recurrent nodes, the entire Transformer model is designed to only require one forward and backward pass through the entire model to process each input sequence. In addition to forming next-step predictive models from these transformer blocks, The authors also provide additional architectural topologies for models tailored to other tasks such as machine translation or text classification, emphasizing the applicability of the Transformer blocks for a diverse array of NLP tasks. Please refer to \cite{Vaswani17} for more information about the composition of a Transformer-based next step predictive mode.
\paragraph{Multi-Head Dot-Product Self-Attention}
\label{sec:attention}
The multi-head dot product self-attention mechanism is the core architectural innovation which enables a Transformer block to fit to temporal correlations present in the training set in one forward and backward pass. On an abstract level, the operation used by the dot product self-attention mechanism with $h$ heads to compute temporal correlations is the the scaled and masked outer product of each $k_i, q_i : i \in \{1, \dots, h\}$ (derived from the input tensor $x \in  \mathbb{R}^n \times \mathbb{R}^d$) as shown in Equation~\eqref{eq:mask}:
\begin{align}
    t_i &= \operatorname{softmax}\left(\frac{\operatorname{mask}(k_i q_i^\top)}{\sqrt{\lfloor d / h \rfloor}}\right) \label{eq:mask}
\end{align}
The results $t_i \in \mathbb{R}^n \times \mathbb{R}^n : i \in \{1, \dots, h\}$ then directly capture how the features of each $k_i$ and $q_i$ are correlated over each pair of time steps in the input sequence. Note that the operator $\operatorname{mask}(\cdot)$ in Equation~\eqref{eq:mask} zeros out lower triangular entries in $k_i q_i^\top \in \mathbb{R}^n \times \mathbb{R}^n$, corresponding to the dot product of features in $q_i$ with features in $k_i$ from previous time steps. Please refer to \cite{Vaswani17} for more information about the multi-head dot product self-attention mechanism and the the Transformer block as a whole.
\subsection{LSTM Enhancements}
This section features two different enhancements featured in Kirill Mavreshko's 
\footnote{Copyright 2018 by Kirill Mavreshko. Source code at \texttt{https://github.com/kpot/keras-transformer}}
implementation of a Transformer-based next-step predictive model that could be independently used with LSTM models to yield performance improvements for student course trajectory modeling in MOOCs. These enhancements have also been independently backed with theoretical justification and empirical experiments, demonstrating performance improvements in coordinated NLP tasks when applied to LSTM models, as further detailed in \cite{Inan16} and \cite{Pereyra17}.
\paragraph{Confidence Penalty Term in Loss}
\label{sec:penalty}
The baseline LSTM model already features some form of regularization, particularly dropout in the weights of the recurrent layers during training. However, for any classification task with a correct output label $y_{true}$ in a set of possible labels $T$, examining the equation for the cross-entropy classification loss
\begin{equation}
L(\theta) = -\log P(y_{true}; \theta)  = -\log p_{true} \quad p_j = P(y_j ; \theta) \ \forall j \in T
\end{equation}
suggests an additional regularization term that penalizes highly confident distributions to reduce overfitting. The confidence penalty uses $H(p(y; \theta))$ as quantitative measure of confidence in a model's output distribution, where a higher value represents a lower level of confidence that the model predicts for each outcome $j \in T$. As a result, the new loss function expands to
\begin{equation}
L^*(\theta) = -\log p_{true} -\beta H(p(y; \theta)) = -\log p_{true} +\beta \sum_{j\in T} p_j \log p_j
\end{equation}
where $\beta$ is a scalar hyperparameter weight for the Confidence Penalty loss term. For theoretical arguments and empirical evidence for adding a confidence penalty term, please refer to \cite{Pereyra17}.
\paragraph{Tied Embedding Layers}
\label{sec:tied}
Another opportunity to introduce additional regularization to any sort of discrete next-step predictive model is found when examining the model's embedding and output layers, specifically
\begin{itemize}
    \item The embedding layer $L$ with dimension $|T| \times d_{embed}$, mapping input tokens in $T$ to vectors in a latent feature space.
    \item The output layer $W$ with dimension $d_{final} \times |T|$, mapping the final intermediate layer output $h$ to an probability logit over the set of all input tokens $T$.
\end{itemize}
After enforcing the condition $d_{embed} = d_{final}$ by inserting a feed-forward layer between the rest of the model and the output later, $W$ is tied to the embedding layer $L$ fixing $W = L^\top$. For theoretical arguments and empirical evidence for the effectiveness of tying the output layer in this fashion, please refer to \cite{Inan16}.
\section{Experiments and Empirical Results}
\label{sec:trajectory}
\subsection{Dataset Cleaning and Processing}
\label{sec:dataset}
\subsubsection{Procedures for Dataset Cleaning and Processing}
Given that the task of student course trajectory modeling requires predicting where a student will navigate next given the student's previous navigation patterns, extensive processing of raw MOOC server logs is required before any training can occur. This process is explained in great detail in \cite{Pardos17}, with the main steps listed below:
\begin{enumerate}
\item Given the raw server log records, select the \texttt{basic\_action} column, timestamp, username, and title columns necessary to build unique course node tokens in step 3.
\item Filter out all log records except those with \texttt{basic\_action} label \texttt{seq\_next}, \texttt{seq\_prev}, or \texttt{seq\_goto}, representing the full set of navigation actions a student can take for each of the MOOCs.
\item Construct a unique positive integer token ID for each course node through concatenating each component of the full course path to construct a unique name for each course node, then assigning each unique name to the token ID.
\item Assemble the full sequence of navigation records for each user by grouping by user ID, then ordering within in each group by timestamp.
\item Prepend the token ID representing the course homepage to every sequence that does not already begin with this token ID, then pad or truncate of the resulting sequences to the maximum sequence length, adding \texttt{0} tokens if necessary.
\end{enumerate}
\subsubsection{Additional Notes on Dataset Selection} In \cite{Pardos17}, additional criteria are included for selecting courses used to demonstrate the utility of a student course trajectory model, including approximating of the entropy of each dataset as a set of discrete random processes via fitting a HMM (Hidden Markov Model) to each dataset. For the experiments in this paper, limited access to MOOC trajectory records preempts the utility of filtering out datasets with low entropy over all course sequences. Table~\ref{tab:dataset} provides summary statistics for the six courses chosen for this paper, hailing from the MOOC offerings of these two universities:
\begin{itemize}
    \item DelftX from the Delft University of Technology in Delft, Netherlands
    \item UCBX from the University of California, Berkeley in Berkeley, California
\end{itemize}
\begin{table}
  \caption{MOOC Course Trajectory Dataset Summary Statistics}
  \label{tab:dataset}
  \begin{tabular}{llccc}
    \hline
    Institution & Course & Term & Nodes & Users\\
    \hline
    DelftX & AE1110X & Fa. 2015 & 291 & 14496\\
    UCBX & EE40LX & Fa. 2015 & 287 & 30633\\
    UCBX & Fin101X & Sp. 2016 & 114 & 2951\\
    UCBX & ColWri2.2X & Sp. 2016 & 54 & 40698\\
    UCBX & CS169.2X & Sp. 2016 & 204 & 940\\
    UCBX & Policy01X & Sp. 2016 & 129 & 1804\\
    \hline
  \end{tabular}
\end{table}

\subsection{Hyperparameters and Training Context}
\label{sec:hyperparameters}

\begin{table}
    \caption{Main Hyperparameters by Architecture Type}
    \label{tab:hyp}
    \begin{tabular}{lcc}
        \hline
        Architecture Type & LSTM & Transformer\\
        \hline
        Max. Seq. Length & 256 & 256\\
        Main Layer Width & 128 & 128\\
        Layer/Block Count & 2 & 2\\
        Attention Heads & N/A & 8\\
        \hline
        Optimizer & Adam & Adam \\
        Learning Rate & 0.01 & 0.0005\\
        Batch Size & 128 & 64\\
        \hline
    \end{tabular}
\end{table}

All training and evaluation was completed on a remote Linux server CPU equipped with 2 GeForce Titan X GPUs (Graphics Processing Units). The script for training and evaluation is written in Python 3 using the Keras \cite{Chollet15} deep learning API over a Tensorflow backend. Table~\ref{tab:hyp} provides the full set of hyperparameters used for training and evaluating each model on each course record dataset. As the goal of this paper is to demonstrate specific differences in model architecture and training that lead to performance gains relative to the earlier results, hyperparameter tuning was not done for any of the LSTM models to facilitate comparison with results in \cite{Pardos19}. Additionally, minimal hyperparameter tuning was done on for the Transformer models in order to minimize the risk of overfitting to the datasets for each course.
\paragraph{Simultaneous Fitting to Multiple Datasets} Since records from each of the 6 courses were processed as described in Section~\ref{sec:dataset} independently, attempting to fit models on multiple courses would result in collisions between different sets of course node tokens. Nonetheless, building a predictive model that can fit to datasets from a wide range of courses is a well-defined area for future research.
\subsection{Empirical Results}
Table~\ref{tab:lstm} presents summary statistics for each model's final test accuracy and total training time per batch for each of the six datasets listed in Table~\ref{tab:dataset}. Table~\ref{tab:transformer} presents additional metrics for the Transformer model pertinent to the analysis in Section~\ref{sec:overfitting}. All statistics in both Table~\ref{tab:lstm} and Table~\ref{tab:transformer} are recorded using the default set of Keras command line logging tools.
\footnote{BaseLogger and ProbarLogger Callback utilities. \cite{Chollet15}}
\begin{table*}
 \caption{Overall Performance Metrics by Architecture}
 \label{tab:lstm}
  \begin{tabular}{lccccc}
    \hline
    Model & Baseline LSTM & LSTM w/ Conf. Penalty & LSTM w/ Tied Emb. & LSTM w/ Both Enh. & Transformer\\
    \hline
    & \multicolumn{5}{c}{Final Test Accuracy}\\
    \hline
    Average & 0.6373 & 0.6355 & 0.6418 & 0.6388 & 0.6383 \\
    Std. Dev. & 0.05623 & 0.05558 & 0.05728 & 0.05592 & 0.05455 \\
    \hline
    & \multicolumn{5}{c}{Training Time per Batch}\\
    \hline
    Average & 35 ms & 35 ms & 35 ms & 35 ms & 2 ms \\
    Std Dev. & 0.94 ms & 0.92 ms & 0.89 ms & 0.86 ms & 0.04 ms \\
    \hline
  \end{tabular}
\end{table*}
\begin{table}
 \caption{Additional Performance Metrics for the Transformer Model}
 \label{tab:transformer}
  \begin{tabular}{lcc}
    \hline
     & Test Acc. & Train Acc. \\
    \hline
    Average & 0.6307 & 0.6383\\
    Standard Deviation & 0.06081 & 0.05455\\
    Avg. for Large Datasets & 0.6438 & 0.6411\\
    Avg. for Small Datasets & 0.6175 & 0.6355\\
    \hline
  \end{tabular}
\end{table}
\subsection{Analysis and Further Considerations}
\subsubsection{Baseline LSTM Comparison with Previous Results}
At face value, the results for the Baseline LSTM model corroborate those presented in \cite{Pardos17}, with the caveat that average accuracy metrics reported in \cite{Pardos17} are calculated in a different fashion that effectively gives more weight to correctly predicting tokens that occur in shorter course trajectory sequences.
\subsubsection{Comparison of Final Test Accuracy Between Models}
Table~\ref{tab:lstm} and Table~\ref{tab:transformer} show that the Transformer model achieves an average final test accuracy of around 63 percent, approximately on par with the average final test accuracy of both LSTM models without tied embeddings, in contrast with a marginally yet consistently higher 64 percent average for the LSTM models that use Tied Embeddings (as described in Section~\ref{sec:tied}). On the other hand, including the Confidence Penalty (as described in Section~\ref{sec:penalty}) does not provide any meaningful improvement in final test accuracy for any of the six datasets. As the training scheme for all models featured in this paper invoke early stopping after 3 epochs without improving validation loss, training any of the above models for more epochs will most likely lead to overfitting on the training set. \subsubsection{Further Analysis of Final Test Accuracy for Transformer Models}
\label{sec:overfitting}
At a first glance, the final test accuracy results in Table~\ref{tab:transformer} seem to contradict Transformer models' considerable performance improvements over LSTM models demonstrated in \cite{Vaswani17}. Nevertheless, the largest dataset featured in this paper only includes 40,698 course trajectory sequences, which is multiple orders of magnitude smaller than the WMT machine translation datasets used in \cite{Vaswani17} with millions of sentence pairs per language pair. This discrepancy in dataset size can cause overfitting for a particular deep learning architecture optimized to train with much larger datasets, even while controlling for model and training hyperparameters. Furthermore, the final training accuracies listed in Table~\ref{tab:transformer} suggest that the Transformer model has overfit to the smaller datasets in this paper despite the use of early stopping, particularly for the following two datasets from courses with fewer than 2,000 unique users as recorded in Table~\ref{tab:dataset}:
\begin{itemize}
\item UCBX CS169.2X with 904 unique users
\item UCBX Policy01X with 1,804 unique users
\end{itemize}
In conclusion, these results provide evidence that Transformer-based models do not yield benefits in accuracy over LSTM models when trained with datasets of similar size to the MOOC course trajectory datasets featured in this paper, in contrast with the much larger datasets common to certain NLP tasks such as machine translation.
\subsubsection{Further Analysis of LSTM Enhancements}
Given that both the Tied Embedding Layers and the Confidence Penalty are theoretically motivated by a search for new forms of model regularization, the empirical results in Table~\ref{tab:lstm} indicate the Tied Embeddings are a marginally more effective form of regularization than the Confidence Penalty for this task. Furthermore, since the Tied Embedding enhancement specifically targets the input embedding and output layers of a discrete next-step predictive model for regularization in contrast to the Confidence Penalty altering the entire model's loss function, the embedding and output layers of each of the LSTM models play a disproportionately important role in the model's performance as a whole for this task.
\subsubsection{Analysis of Training Time Results}
In contrast to the Transformer model's lack of improvement in final test accuracies for all six datasets, the results in Table~\ref{tab:lstm} and Table~\ref{tab:transformer} suggest that the Transformer model outperforms all types of LSTM models by more than an order of magnitude with respect to total training time per batch. Given that both the LSTM and Transformer models are built to accommodate a maximum sequence length of 256 as indicated in Table~\ref{tab:hyp}, the results in Table~\ref{tab:transformer} are consistent with the reduced number of training passes through the model's computational graph per input sequence, as encapsulated in the Transformer architecture's design goals from Section~\ref{sec:transformer}.
\subsection{Directions for Future Research}
\label{sec:future}
\subsubsection{Task-Specific Model Enhancements}
As mentioned in multiple sections of this paper, certain task-specific strategies for improving performance on MOOC course trajectory prediction covered in \cite{Pardos17} are not investigated in this paper, even if applying these strategies in conjunction impact performance in a noteworthy manner. Some of these additional strategies used to improve performance on these two tasks include:
\begin{itemize}
\item Calculating the entropy of each dataset's best-fit HMM transition matrix as a criterion for selecting MOOC course trajectory datasets used to evaluate the enhanced LSTM and Transformer models.
\item Incorporating auxiliary data inputs, including the time difference between course navigation actions, into evaluating the benefits of enhanced LSTM models and Transformer models over baseline results from \cite{Pardos17}.
\end{itemize}
\subsubsection{Model Pre-Training and Multitask Learning}
Another more ambitious goal for further research involves constructing one model that can provide meaningful predictions for multiple tasks with minimal training needs. Given the wide applicability of models with generalized predictive modeling capabilities, this model would most likely incorporate innovations originally designed to provide multitask capabilities for NLP applications. For example, \cite{Radford19} presents a NLP model that is first trained to perform a next-word prediction task on large text datasets before undergoing fine-tune training for more specific downstream NLP tasks, which include tasks such as text classification, sentence embedding, question answering and free-form text generation. In the context of education analytics tasks, an analogous suite of tasks for such a model could include modeling overall course performance and individualized suggestions for instructors assisting students with course material.



%
\bibliographystyle{abbrv}
\bibliography{preprint}  
%
%
\balancecolumns
\end{document}